\documentclass[sigconf,nonacm]{acmart}
\AtBeginDocument{%
  }

\setcopyright{none}
\acmConference[KDD '27]
  {ACM SIGKDD Conference on Knowledge Discovery and Data Mining}
  {August 1--5, 2027}
  {San Jose, CA, USA}

\acmDOI{}
\acmISBN{}

\usepackage{tikz}
\usepackage{tikz}
\usetikzlibrary{positioning, arrows.meta, shapes.geometric, backgrounds, fit}
\usepackage{comment} 
\usepackage{amsthm}
\DeclareMathOperator{\sign}{sign}

\begin{document}

\title{Learning Who Benefits Most:\\ Multi-Treatment Causal Ranking for Population Health Stratification}


\author{Rocco Zaccagnino}
\authornote{Corresponding author}
\email{rzaccagnino@unisa.it}

\author{Martina Girolamo}
\email{m.girolamo1@studenti.unisa.it}

\author{Delfina Malandrino}
\email{dmalandrino@unisa.it}

\author{Alessia Ture}
\email{ature@unisa.it}

\affiliation{%
  \institution{University of Salerno}
  \city{Salerno}
  \country{Italy}
}


\author{Alberto Lombardi}
\email{Alberto.Lombardi@aslbenevento.it}
\affiliation{
  \institution{Local Health Authority of Benevento}
  \city{Benevento}
  \country{Italy}
}


\author{Donatello Telesca}
\email{dtelesca@g.ucla.edu}
\orcid{0000-0003-0773-451X}
\affiliation{%
  \institution{University of California, Los Angeles}
  \city{Los Angeles}
  \country{USA}
}

\renewcommand{\shortauthors}{Zaccagnino et al.}

\begin{abstract}
Population \textit{health stratification} frameworks, such as those defined by the Italian DM77, organize patients according to their level of clinical need and associated care pathways. However, they remain largely descriptive and provide limited practical guidance: although they distinguish among different levels of complexity, they do not indicate which changes in care should be prioritized when healthcare resources are constrained.

We present \texttt{PROMETHEUS} (\textit{Proactive Resource Optimization through Multi-treatment Effect-based Targeting in Healthcare Environments Using Stratification}), a framework that reformulates population health stratification as a \textit{multi-treatment causal ranking problem}. Under capacity constraints, the main objective is not to obtain a perfectly calibrated treatment-effect estimate for every patient, but to preserve the relative ordering of care opportunities according to their expected outcome. \texttt{PROMETHEUS} therefore focuses on identifying which patient-care treatments should be prioritized over others.

In a  single-treatment setting, each treatment, i.e., a specific level of care, would be studied as an independent binary decision: whether or not to intensify care for patients eligible for that specific treatment. In \texttt{PROMETHEUS}, instead, each clinically admissible treatment is treated separately, and all treatments across the stratification pathway are learned jointly. This multi-treatment formulation allows eligible patient-treatment opportunities to be compared on a common scale of expected clinical benefit rather than ranked only within a single treatment. For each treatment, the framework estimates the expected benefit, i.e., the \textit{outcome}, of moving a patient to a more intensive care profile rather than maintaining the current pathway. This benefit can be measured as avoidable hospital admissions, emergency-department use, disease progression, or a composite clinical measure. 

The proposed model ranks patients according to their relative causal outcomes using a Siamese contrastive neural network. The network learns similar representations for treatments associated with comparable causal effects, while separating those that yield substantially different benefits. This formulation facilitates information sharing across related care transitions and produces a consistent priority ranking across patients and levels of care.

Our results show that the resulting population stratification preserves alignment with the levels and care pathways defined by DM77, while extending beyond conventional descriptive baselines, such as standard comorbidity indices and predictive risk models. Rather than merely describing clinical complexity or forecasting future utilization, \texttt{PROMETHEUS} makes stratification actionable by identifying which patients should receive a more intensive level of care first, with the goal of maximizing population-level benefit under limited capacity.
\end{abstract}



\maketitle

\section{Introduction}

Population aging, multimorbidity, and workforce shortages are increasing pressure on health and long-term care systems \cite{oecd2024health,barnett2012epidemiology,chowdhury2023global}. Patients with multiple chronic conditions often require coordinated and resource-intensive services, such as case management, home monitoring, specialist coordination, and transitional care \cite{moffat2015challenges}. Because these services cannot be delivered to the entire population with the same intensity \cite{stokes2015effectiveness}, healthcare systems must determine not only which needs exist, but also how limited care capacity should be distributed across eligible patients.

Population health stratification supports this task by grouping individuals according to morbidity, vulnerability, clinical complexity, and expected healthcare utilization, linking these profiles to differentiated, tiered care pathways \cite{chong2019segmentation,cerezo2025unlocking}. This shift toward proactive, community-based care is a global priority, operationalized through frameworks such as Accountable Care Organizations  and Patient-Centered Medical Homes in the United States, or Integrated Care Systems within the UK National Health Service. In Italy, this international paradigm is formally codified by Ministerial Decree No.77/2022 (DM77), which establishes the regulatory framework for territorial healthcare. DM77 establishes population stratification as a core organizational principle for identifying heterogeneous clinical needs and structuring progressively differentiated levels of care \cite{ministerosalute2022dm77}, making it the primary operational testbed for our framework.

Structural frameworks are routinely complemented by descriptive case-mix and predictive population-stratification systems. An example is the Johns Hopkins Adjusted Clinical Groups (ACG) system, which characterizes individuals according to their longitudinal morbidity patterns and expected healthcare needs and resource utilization. Other widely used approaches include standard clinical metrics, such as the Charlson and Elixhauser comorbidity indices, the CMS Hierarchical Condition Categories, and machine-learning risk scores, which use demographic, diagnostic, pharmacy, and utilization data to estimate morbidity burden, future expenditure, or adverse-outcome risk \cite{weiner1991development,charlson1987new,elixhauser1998comorbidity,ash2000risk}. These descriptive systems are valuable for identifying clinically complex or high-risk patients, but capacity-constrained allocation requires answering a fundamentally different question: \textit{which eligible patients are expected to benefit the most from a more intensive care profile?}

Predicted risk and expected treatment benefit are distinct. A patient at high risk of hospitalization may have a largely non-modifiable trajectory and derive limited additional benefit from intensive case management, whereas a patient at moderate risk may respond more favorably to early treatment, remote monitoring, or transitional care \cite{kansagara2011risk,marafino2020preventable}. Risk-based allocation may therefore conflate \textit{clinical need}, reflecting disease burden and essential care requirements, with \textit{clinical actionability}, reflecting the expected benefit of a specific treatment \cite{lewis2010impactibility,orlowski2021impactibility}.

We adopt a \textit{causal} perspective in which each clinically admissible care profile represents a distinct treatment option. For an eligible patient, treatment consists of assignment to a specific profile that is more intensive than the current care pathway, while maintenance of the current pathway serves as the comparison condition. Because a patient may be eligible for multiple profiles, the same patient can generate several candidate treatment opportunities, each corresponding to a different potential transition in care.

The \textit{outcome} is defined as a health event measured over a fixed follow-up horizon, such as 6 or 12 months. Relevant outcomes may include hospitalization, emergency department use, mortality, disease progression, or healthcare utilization. The outcome is oriented so that larger values consistently represent greater benefit, for example through a reduction in hospitalizations, and is defined on a common clinical scale across all candidate treatments. This common definition is necessary for comparing the expected benefits of different care-profile assignments.

Rather than seeking precisely calibrated treatment-effect estimates for every patient and every profile, our objective is to determine which patient--treatment opportunities are expected to produce relatively larger improvements compared with maintenance of the current pathway.

Under strict capacity constraints, the key operational requirement is often the relative ordering of these opportunities: patient--treatment pairs associated with larger expected benefits should receive priority over those associated with smaller benefits. This motivates \textit{causal ranking}, which aims to preserve the ordering induced by heterogeneous treatment effects without requiring perfectly calibrated individual effect estimates \cite{arno2026rank}.

Most existing causal ranking formulations consider a single binary treatment and therefore rank patients according to their expected benefit from one intervention. Population health stratification presents a more general setting: several clinically admissible care profiles may be available, and both patients and treatment options must be prioritized. Training an independent ranking model for each profile would produce treatment-specific orderings whose scores are not necessarily comparable, even when the corresponding services compete for the same financial, technological, or workforce resources. We therefore formulate population health stratification as a \textit{multi-treatment causal ranking problem}, in which eligible patient--profile assignments are jointly ordered according to their expected incremental benefit.

The unit of prioritization is consequently a \textit{patient--treatment opportunity}: the possibility of assigning an eligible patient to a specific care profile. Because all opportunities are evaluated with respect to a common outcome, follow-up horizon, direction of benefit, and measurement scale, their treatment-specific effects can be represented within a shared \textit{global ordering}. This enables prioritization both among patients competing for the same profile and among different profiles competing for shared resources.

To this aim, we introduce \texttt{PROMETHEUS}, a framework for learning a globally comparable ordering of patient--profile opportunities. Its causal supervision is constructed separately for each candidate care profile through cross-fitted doubly robust pseudo-outcomes. For an eligible patient (i) and profile (p), this pseudo-outcome provides a noisy estimate of the incremental benefit of assigning the patient to (p), relative to maintaining the current care pathway. These quantities are not interpreted as final individualized treatment-effect estimates. Instead, they define the causal signal on which the ranking problem is based: when the estimated benefit associated with one opportunity is sufficiently larger and more reliable than that associated with another, the corresponding pair provides evidence about their relative priority.

\texttt{PROMETHEUS} learns this ordering through a  Siamese ranking network. Two patient-profile opportunities are processed by identical branches with shared parameters, producing scores on a common scale. Their score difference is trained to agree with the ordering induced by the doubly robust causal signals. 


The shared representation can be initialized through contrastive pretraining. During this stage, opportunities associated with similar and sufficiently stable causal signals are encouraged to have nearby latent representations, whereas opportunities with clearly different expected benefits are separated. This initialization gives the Siamese ranker a representation organized around causal-response similarity before the direct pairwise-ranking objective is optimized. 


We evaluate \texttt{PROMETHEUS} through a controlled synthetic benchmark and a semi-synthetic study built on pseudonymized administrative data from ASL
Benevento. The synthetic benchmark provides complete causal and operational
ground truth. The ASL study preserves real population, diagnosis, and
utilization structure while simulating assignments and outcomes. Both settings
are preclinical: they test the method and its failure modes, not the real
clinical effectiveness of DM77-aligned services.

To promote transparency and reproducibility, we provide the source code for \texttt{PROMETHEUS}, including the data-processing pipeline, causal-estimation workflow, model implementation, and evaluation scripts.\footnote{\url{https://github.com/FLaTNNBio/PROMETHEUS}}

\section{Related Work}

\texttt{PROMETHEUS} connects population-health stratification, heterogeneous
treatment-effect estimation, learning to rank, and constrained allocation.
DM77-aligned levels, case-mix systems, comorbidity indices, and risk models
describe need or prognosis, but they do not estimate which candidate service
is most beneficial for the same patient
\cite{ministerosalute2022dm77,elixhauser1998comorbidity,ash2000risk}.
Recent work has also explored causal-temporal representation learning for
modeling comorbidity progression risk from longitudinal primary-care data
\cite{zaccagnino2026causal}. Such approaches improve the representation of
patient trajectories and disease progression, but address a prognostic
objective rather than the comparative prioritization of clinically admissible
patient--profile opportunities. Historical risk-based allocation can also
reproduce unequal access when past utilization reflects barriers rather than
need \cite{obermeyer2019dissecting}.

Causal forests, meta-learners, orthogonal estimators, and neural causal models
estimate heterogeneous responses to interventions
\cite{athey2019generalized,kunzel2019metalearners,kennedy2023optimal,
nie2021quasi,johansson2016learning,shalit2017estimating,shi2019adapting}.
Multi-task models share information across related interventions
\cite{louizos2017causal,alaa2017bayesian}. In contrast to methods centered on
numerical effect estimation, our target is an ordinal priority score over
patient--profile opportunities.

Uplift modeling and causal ranking methods directly optimize the ordering of
intervention benefit
\cite{devriendt2020learning,yadlowsky2023rate,kamran2024learning,
vanderschueren2024metalearners,arno2026rank}. Most consider one intervention
contrast at a time. Policy-learning methods can optimize multi-action
assignments under a fixed objective and resource configuration
\cite{athey2021policy,kallus2018confounding,zhou2023offline}.
\texttt{PROMETHEUS} instead learns one reusable ordering across comparable
profile-specific contrasts and applies the operational constraints afterward.

Contrastive causal methods have mainly been used to improve balance, outcome
prediction, or treatment-effect representations
\cite{zhu2024contrastive,zaccagnino2026hermes}. Here, contrastive learning
serves a different role: it provides an initial representation before direct
ranking.

\section{Problem Formulation}
\label{sec:problem-formulation}
We formalize population-health stratification as a multi-treatment causal ranking problem. The framework separates three questions that are often combined in conventional stratification systems: how much clinical and care support a patient needs, which admissible care profile is expected to provide the largest incremental benefit, and which recommended services can be activated with the available resources. The formulation proceeds in two steps. We first define the patient information, care profiles, eligibility rules, and profile-specific causal targets. We then define the global ranking objective and show how the learned ordering is converted into patient-level recommendations and capacity-constrained allocations. Throughout the formulation, clinical eligibility remains governed by external rules, while the learned model is used only to prioritize opportunities that are already considered clinically admissible
\subsection{Decision Space and Causal Targets}
\label{sec:decision-space}
We consider a population-health planning cycle at a predefined index date where the dataset can be written as $\mathcal D = \left\{ X_i,\, C_i,\, A_i,\, Y_i \right\}_{i=1}^{n}$. For
patient $i$, let $X_i$ denote the demographic, clinical, functional,
pharmaceutical, social, and prior-utilization  covariates, and let $C_i$ denote the care pathway currently received. The
observed treatment is $A_i\in{0}\cup\mathcal P$, where $0$ denotes
maintenance of the current pathway and
$\mathcal P={1,\ldots,M}$ is the set of candidate care treatments. These treatments may include, for example, remote monitoring, transitional care, home-based support, or multidisciplinary case management. The outcome $Y_i$ is
measured over a prespecified follow-up horizon and is defined so that larger values represent a more favorable clinical result.

Different treatments can be included in the same global ranking only when their expected benefits have the same clinical interpretation. In particular, they must be evaluated against the same reference treatment and use the same
outcome definition, follow-up horizon, measurement unit, and direction of
benefit. Treatments that pursue substantially different clinical objectives
belong to separate comparability domains and are not ranked against one
another. The formal comparability conditions are reported in
Appendix~\ref{app:technical-details}.

A separate descriptive module produces a baseline need level,
$
\widehat L_i^{\mathrm{need}} = f_\phi(X_i^{\mathrm{need}}),
$
where $f_\phi$ may be a rule-based system, an ordinal model, an existing stratification tool, or a hybrid procedure. The resulting level summarizes the patient’s clinical and care complexity and is retained as a distinct output for population stratification and reporting. It may also be used as contextual information by the downstream causal model, but it neither estimates treatment benefit nor determines which profile should be prioritized. Current care is excluded as a direct predictor to avoid inferring need merely from the service already received.

Before a treatment can be ranked, the framework determines whether it is
clinically admissible for the patient. For each candidate treatment $p$, let
$g_p$ denote the set of externally defined clinical and organizational rules
associated with that treatment. These rules  consider the patient's
pre-decision characteristics $X_i$, the current care pathway $C_i$, and the
estimated baseline need level $\widehat L_i^{\mathrm{need}}$.

The admissibility indicator is
$
E_i^{(p)}
=
g_p
\left(
X_i,
C_i,
\widehat L_i^{\mathrm{need}}
\right),
$
where
\[
E_i^{(p)}
=
\begin{cases}
1, & \text{if treatment $p$ may be considered for patient $i$},\\
0, & \text{otherwise}.
\end{cases}
\]
The set
$
\mathcal P_i^{\mathrm{adm}}
=
\left\{
p\in\mathcal P:
E_i^{(p)}=1
\right\}
$
therefore contains all candidate treatments that satisfy the required
eligibility criteria for patient $i$.

For example, remote monitoring may require a chronic condition suitable for
home monitoring, while transitional care may require a recent hospitalization
or discharge. A treatment may also be excluded because of a
contraindication, an unmet prerequisite, or incompatibility with the current
care pathway.

The rules represented by $g_p$ are specified outside the ranking model by
clinical and organizational governance. The model does not learn whether a
patient is eligible and does not compare treatments that have already been
excluded by these rules. It only assigns priority scores to the admissible
patient--treatment opportunities. Consequently, a high ranking score cannot
create eligibility, remove a contraindication, or replace professional
judgment.

The decision unit is a \emph{patient--profile opportunity}
$u=(i,p)$ with $p\in\mathcal P_i^{\mathrm{adm}}$. One patient may generate
several opportunities because several services may be clinically admissible.
These opportunities can differ in expected benefit, cost, and resource use.
For this reason, one generic patient-level risk score cannot fully describe the
decision.
 Let
$D_i^{(p)}=\mathbb I(A_i=p)$ in the eligible population receiving either
profile $p$ or maintenance, and let
$W_i=(X_i,C_i,\widehat L_i^{\mathrm{need}})$. The profile-specific target is
\begin{equation}
\tau_p(w)
=
\mathbb E
\left[
Y_i^{(p)}-Y_i^{(0)}
\mid
W_i=w,\,
E_i^{(p)}=1
\right].
\label{eq:profile-benefit}
\end{equation}
It is the expected improvement from activating profile $p$ rather than
maintaining the current pathway for eligible patients with state $w$.
Identification requires profile-specific consistency, conditional
exchangeability, positivity, and an appropriate interference assumption.
Clinical eligibility alone does not guarantee that the contrast is supported
by the observed data. Appendix~\ref{app:technical-details} gives the complete
identification conditions.

\subsection{Global Ranking, Recommendation, and Allocation}
\label{sec:global-target}

The learning target is a shared score $s_{i,p}$ over admissible and supported
opportunities. For comparable opportunities $u=(i,p)$ and $v=(j,q)$, the
desired ordering is
\begin{equation}
\tau_p(W_i)>\tau_q(W_j)
\quad\Longrightarrow\quad
s_{i,p}>s_{j,q}.
\label{eq:global-ordering}
\end{equation}

The term \emph{global} refers to three comparison domains. \textit{Within-profile} comparisons rank patients competing for the same service; \textit{within-patient} comparisons rank alternative admissible services for the same patient; and \textit{cross-profile} comparisons rank opportunities involving different patients and different services. Cross-profile comparisons connect the profile-specific orderings into a single population-level priority axis.

Cross-profile supervision is used only when the profiles belong to the same comparability domain and both opportunities have adequate empirical support. Without this condition, separate profile-specific rankings would not provide a valid basis for allocating shared resources.

The ranker assigns a priority score $s_{i,p}$ to each admissible
patient--treatment opportunity. This score is initially used only to order
the opportunities: a higher value indicates that treatment $p$ is expected
to provide a larger benefit for patient $i$ than an opportunity receiving a
lower score. However, the numerical value of the score has no direct clinical
interpretation. 

A separate calibration step is therefore applied on an independent calibration
set:
$
\widehat b_{i,p}=h(s_{i,p}),
$
where $h_p$ is a monotone function fitted for treatment $p$. Monotonicity
ensures that the ordering learned by the ranker is preserved: if one
opportunity has a higher score than another, calibration cannot reverse their
priority. The calibrated quantity $\widehat b_{i,p}$ provides an approximate
benefit scale that can be compared with the minimum-benefit threshold and used
by the allocation optimizer. It remains a model-based estimate under the
causal assumptions and is not presented as an observed or precisely identified
individual treatment effect.

For each patient, the recommendation layer selects the supported admissible
profile with the largest calibrated benefit above a minimum threshold. When no
profile exceeds the threshold, the current pathway is maintained. When no
profile has adequate empirical support, the model abstains and sends the case
to professional review.

The allocation layer then chooses among positive recommendations under the
available capacities. Let $Z_{i,p}=1$ indicate that recommended profile $p$ is
activated for patient $i$. The objective is
\begin{equation}
\max_Z
\sum_i
\sum_{p\in\mathcal P_i^{\mathrm{sup}}}
Z_{i,p}\widehat b_{i,p},
\label{eq:allocation-target}
\end{equation}
subject to recommendation consistency, mutual exclusivity, profile capacity,
shared resources, budget, geography, and organizational rules. A recommended
profile may therefore remain unallocated when capacity is unavailable. The
patient is recorded as operationally deferred; the need level, eligibility,
and recommendation do not change. Appendix~\ref{app:technical-details} gives
the compact formal rules.

\section{PROMETHEUS Framework}
\label{sec:prometheus}
\texttt{PROMETHEUS} uses profile-specific causal signals to learn a priority ordering over clinically admissible patient--profile opportunities.

\begin{figure*}[t]
    \centering
    \resizebox{0.92\linewidth}{!}{%
    \begin{tikzpicture}[
        font=\sffamily,
        >=Stealth,
        tensor/.style={rectangle, draw=black!70, fill=gray!10, thick, minimum width=2.5cm, minimum height=1cm, align=center},
        encoder/.style={rectangle, rounded corners, draw=blue!70, fill=blue!5, thick, minimum width=2cm, minimum height=1.2cm, align=center},
        latent/.style={circle, draw=purple!70, fill=purple!5, thick, minimum size=1.2cm, align=center},
        head/.style={rectangle, rounded corners, draw=teal!70, fill=teal!5, thick, minimum width=1.8cm, minimum height=1cm, align=center},
        score/.style={rectangle, draw=black!70, fill=white, thick, minimum width=1.2cm, minimum height=0.8cm, align=center, font=\bfseries},
        lossbox/.style={rectangle, rounded corners, draw=red!70, fill=red!5, thick, minimum width=3.5cm, minimum height=1.2cm, align=center},
        causalbox/.style={rectangle, draw=orange!80, fill=orange!10, dashed, thick, minimum width=3cm, minimum height=1cm, align=center},
        arrow/.style={->, thick, draw=black!70},
        dashedarrow/.style={->, thick, dashed, draw=black!60}
    ]

    \node[tensor] (input1) {Patient $W_i$ \\ Profile $e_p^{\mathrm{prof}}$};
    \node[encoder, right=1cm of input1] (enc1) {Shared\\Encoder $F_\theta$};
    \node[latent, right=1.2cm of enc1] (h1) {$h_{i,p}$};
    \node[head, right=1.2cm of h1] (rank1) {Rank\\Head $r_\theta$};
    \node[score, right=1cm of rank1] (s1) {$s_{i,p}$};

    \draw[arrow] (input1) -- (enc1);
    \draw[arrow] (enc1) -- (h1);
    \draw[arrow] (h1) -- (rank1);
    \draw[arrow] (rank1) -- (s1);

    \node[tensor, below=2cm of input1] (input2) {Patient $W_j$ \\ Profile $e_q^{\mathrm{prof}}$};
    \node[encoder, right=1cm of input2] (enc2) {Shared\\Encoder $F_\theta$};
    \node[latent, right=1.2cm of enc2] (h2) {$h_{j,q}$};
    \node[head, right=1.2cm of h2] (rank2) {Rank\\Head $r_\theta$};
    \node[score, right=1cm of rank2] (s2) {$s_{j,q}$};

    \draw[arrow] (input2) -- (enc2);
    \draw[arrow] (enc2) -- (h2);
    \draw[arrow] (h2) -- (rank2);
    \draw[arrow] (rank2) -- (s2);

    \draw[<->, thick, draw=purple!80, dashed] (h1) -- node[fill=white, align=center, font=\footnotesize, inner sep=2pt] {Contrastive\\Regularization} (h2);

    \path (s1.east) -- coordinate (mid) (s2.east);
    \node[lossbox, right=1.5cm of mid] (loss) {Pairwise Ranking Loss\\[4pt] $\mathcal{L}_{\mathrm{rank}}(u,v)$};

    \draw[arrow] (s1.east) -- ++(0.5,0) |- (loss.160);
    \draw[arrow] (s2.east) -- ++(0.5,0) |- (loss.200);

    \node[causalbox, above=1.5cm of loss] (dr) {Doubly Robust (DR)\\Pseudo-outcomes\\[2pt] $\overline{\Gamma}_i^{(p)}, \overline{\Gamma}_j^{(q)}$};
    
    \draw[dashedarrow] (dr) -- node[right, font=\footnotesize, align=left] {Supervision Signal\\$y_{uv} = \sign(\overline{\Gamma}_i^{(p)} - \overline{\Gamma}_j^{(q)})$} (loss);

    \begin{scope}[on background layer]
        \node[rectangle, draw=blue!30, fill=blue!2, dashed, rounded corners, inner sep=0.3cm, fit=(enc1) (enc2) (rank1) (rank2) (h1) (h2) (s1) (s2)] (siamese) {};
        \node[anchor=south west, font=\bfseries\color{blue!80}, xshift=0.2cm, yshift=0.1cm] at (siamese.north west) {Siamese Network};
    \end{scope}

    \end{tikzpicture}
    }
    \caption{The \texttt{PROMETHEUS} multi-treatment causal ranking architecture. Admissible patient-profile opportunities $(u, v)$ are processed by a shared Siamese encoder $F_\theta$. The latent representations $h$ are organized via uncertainty-aware contrastive pretraining based on their estimated causal responses. A ranking head $r_\theta$ outputs priority scores $s$, optimized through a pairwise ranking loss $\mathcal{L}_{\mathrm{rank}}$ supervised by cross-fitted doubly robust pseudo-outcomes.}
    \Description{Sample teaser image showing the main concept of the paper.}
    \label{fig:architecture}
\end{figure*}
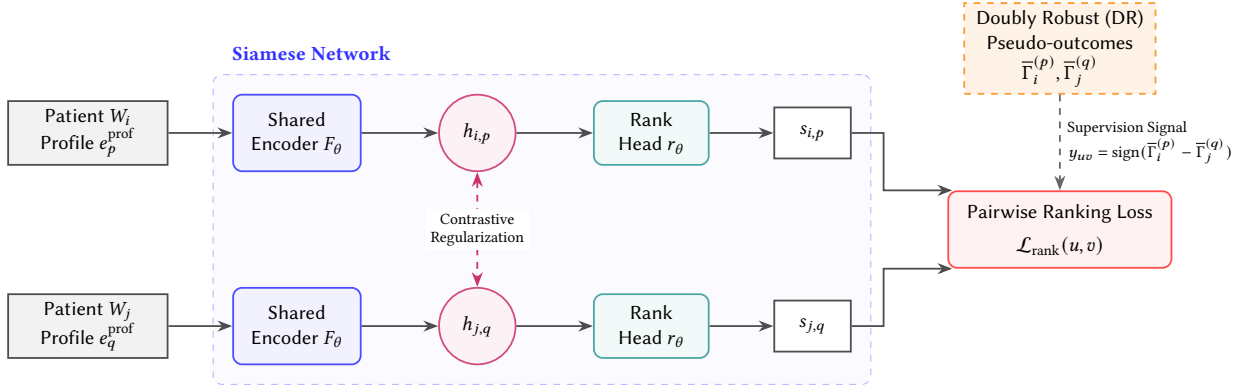

\texttt{PROMETHEUS} first constructs profile-specific doubly robust causal signals through cross-fitted nuisance models. These signals supervise a Siamese ranking network, which assigns comparable priority scores to patient--profile opportunities by learning from pairwise differences in expected benefit. Reliability checks filter or downweight comparisons that are weakly supported or dominated by estimation noise. The resulting scores are then converted into patient-level recommendations and integrated with resource constraints to obtain a feasible allocation. A contrastive pretraining step initializes the shared encoder by organizing opportunities according to the similarity of their causal signals. Figure~\ref{fig:architecture} summarizes this pipeline.

\subsection{Cross-Fitted Causal Supervision}

For profile $p$, let $e_p(w)=\Pr(D_i^{(p)}=1\mid W_i=w,E_i^{(p)}=1)$ be the propensity score and $\mu_{p,d}(w)=\mathbb E[Y_i\mid D_i^{(p)}=d,W_i=w,E_i^{(p)}=1]$ be the outcome regressions. The doubly robust pseudo-outcome is defined as:
\begin{align}
\Gamma_i^{(p)}
={}&
\widehat\mu_{p,1}(W_i)-\widehat\mu_{p,0}(W_i)
\nonumber\\
&+
\frac{D_i^{(p)}[Y_i-\widehat\mu_{p,1}(W_i)]}{\widehat e_p(W_i)}
-
\frac{(1-D_i^{(p)})[Y_i-\widehat\mu_{p,0}(W_i)]}{1-\widehat e_p(W_i)}.
\label{eq:dr}
\end{align}

Nuisance predictions are generated out-of-fold at the patient level to prevent data leakage across profiles. Cross-fitting is repeated over multiple split/initialization runs: the mean $\overline\Gamma_i^{(p)}$ acts as the causal supervision signal. 
Under the assumptions in Appendix~\ref{app:technical-details}, $\mathbb{E}[\Gamma_i^{(p)}]$ equals the target in Eq.~\eqref{eq:profile-benefit}. Individual pseudo-outcomes remain noisy and are not treated as deterministic individual effects.

Additionally, this stage records propensity support, treatment prevalence, effective sample size, residual magnitude, and cross-fitting instability to identify reliable comparisons and candidate recommendations.

\subsection{Reliability-Aware Pair Construction}

The ranking model is trained on pairs of patient--intervention opportunities. To filter out comparisons dominated by nuisance estimation noise, we evaluate the signal difference $d_{uv} = \left|\overline\Gamma_u-\overline\Gamma_v\right|$ against the combined instability $r_{uv} = \sqrt{\sigma_u^2+\sigma_v^2+\epsilon}$. The ratio $z_{uv} = d_{uv} / r_{uv}$ measures signal separation relative to estimation noise.

A pair of opportunities $u=(i,p)$ and $v=(j,q)$ is retained for ranking ($M_{uv}^{\mathrm{pair}}=1$) only when three binary conditions are satisfied:
$$
M_{uv}^{\mathrm{pair}}
=
M_{uv}^{\mathrm{comp}}
M_{uv}^{\mathrm{sup}}
\mathbb I\{z_{uv}\geq\kappa\}.
$$
Here, $M_{uv}^{\mathrm{comp}}=1$ indicates that both opportunities belong to the same comparability domain, $M_{uv}^{\mathrm{sup}}=1$ ensures sufficient overlap, treatment prevalence, and effective sample size, and $\mathbb{I}\{z_{uv}\geq\kappa\}=1$ enforces a standardized separation above threshold $\kappa$.

Pairs with $M_{uv}^{\mathrm{pair}}=0$ are excluded from training. For retained pairs, the ordering target is $y_{uv} = \operatorname{sign}\left(\overline\Gamma_u-\overline\Gamma_v\right)$, indicating which opportunity yields higher estimated benefit. Retained pairs are also assigned a reliability weight $\omega_{uv}$ based on causal separation and overlap metrics, controlling their relative influence in the loss function.

Pair construction relies strictly on the ranker-training partition; pseudo-outcomes from validation, calibration, or test sets are never used. Detailed support criteria and weighting definitions are provided in Appendix~\ref{app:training-details}.

\subsection{Network Architecture and Two-Stage Training}
\label{sec:network-training}

Each opportunity $u=(i,p)$ is input as $x_{i,p} = \left[W_i \;\Vert\; e_p^{\mathrm{int}}\right]$, concatenating pre-decision patient features $W_i=(X_i,C_i,\widehat L_i^{\mathrm{need}})$ with a learned intervention embedding $e_p^{\mathrm{int}}$.

A shared encoder maps this input to a latent vector $h_{i,p} = F_\theta(x_{i,p})$, which a ranking head converts into a scalar priority score $s_{i,p} = r_\psi(h_{i,p})$. Architectural details are reported in Appendix~\ref{app:network-details}.

\paragraph{Stage 1: Contrastive representation pretraining.}

To initialize the encoder, let $\mathcal C^{+}$ contain reliable pairs with similar causal signals, and $\mathcal C^{-}$ contain reliable pairs with clearly distinct signals (unstable pairs are omitted). With latent distance $d_{uv}^{h} = \left\|h_u-h_v\right\|_2$, the pretraining objective is:
\begin{align}
\mathcal L_{\mathrm{pre}}
={}&
\frac{1}{|\mathcal C^{+}|}
\sum_{(u,v)\in\mathcal C^{+}}
\eta_{uv}
\left(d_{uv}^{h}\right)^2
\nonumber\\
&+
\frac{1}{|\mathcal C^{-}|}
\sum_{(u,v)\in\mathcal C^{-}}
\eta_{uv}
\left[m-d_{uv}^{h}\right]_{+}^2
+
\lambda_{\mathrm{pre}}\mathcal R_{\mathrm{pre}}(\theta),
\label{eq:pretraining-loss}
\end{align}
where $m>0$ is the separation margin, $[a]_{+}=\max(0,a)$, and $\eta_{uv}$ is a reliability weight pulling similar representations together and pushing distinct ones apart up to margin $m$.

\paragraph{Stage 2: Global ranking fine-tuning.}

Using the retained pairs and ordering labels $y_{uv}$, the pairwise loss for a pair $(u,v)$ is:
\[
\ell_{\mathrm{rank}}(u,v)
=
\omega_{uv}
\log
\left[
1+
\exp
\left\{
-y_{uv}(s_u-s_v)
\right\}
\right].
\]

Let $\mathcal K = \{\mathrm{within\mbox{-}intervention}, \mathrm{within\mbox{-}patient}, \mathrm{cross\mbox{-}intervention}\}$ denote the pair types. The normalized loss for category $k$ is:
\[
\mathcal L_{\mathrm{rank}}^{(k)}
=
\frac{
\sum_{(u,v)\in\mathcal B_k}
M_{uv}^{\mathrm{pair}}
\ell_{\mathrm{rank}}(u,v)
}{
\sum_{(u,v)\in\mathcal B_k}
M_{uv}^{\mathrm{pair}}
+\epsilon
},
\]
where $\mathcal B_k$ is the set of sampled pairs of type $k$. The complete fine-tuning objective balances all pair types:
\begin{equation}
\mathcal L_{\mathrm{final}}
=
\sum_{k\in\mathcal K}
\alpha_k
\mathcal L_{\mathrm{rank}}^{(k)}
+
\lambda_{\mathrm{reg}}
\mathcal R_{\mathrm{rank}}(\theta,\psi).
\label{eq:final-training-objective}
\end{equation}
Normalizing each category separately ensures that the larger volume of cross-patient pairs does not dominate training.

The overall model optimization follows the sequential scheme:
$
\mathcal L_{\mathrm{pre}} \rightarrow \mathcal L_{\mathrm{final}}
$. After pretraining, $F_\theta$ and $r_\psi$ are fine-tuned using Eq.~\eqref{eq:final-training-objective} alone, as joint optimization yielded no additional benefit in ablation experiments.

\section{Experimental Setup}
\label{sec:experimental-setup}

We evaluate the complete \texttt{PROMETHEUS} pipeline across synthetic and ASL semi-synthetic settings. The synthetic benchmark provides full causal and operational ground truth, whereas the ASL study preserves the heterogeneity of a real healthcare population under locked simulated mechanisms. Oracle quantities are used only for final evaluation after freezing the non-oracle pipeline.

\subsection{Evaluation Settings}
\label{sec:evaluation-settings}

\paragraph{Synthetic Benchmark}
\label{sec:synthetic-benchmark}

The synthetic benchmark generates the complete population-health decision
problem. For each patient, it includes pre-decision covariates, the current care
pathway, a DM77-aligned baseline-need level, intervention-specific eligibility,
non-random treatment assignment, heterogeneous potential outcomes, resource
costs, service capacities, and oracle recommendations and allocations.

Six discretionary care interventions are considered. Each intervention is
evaluated against maintenance of the current pathway using the same outcome:
hospital-free days over a 365-day follow-up period. A protected palliative
pathway is generated separately but excluded from the global ranking because it
pursues a different clinical objective and therefore belongs to a separate
comparability domain. The benchmark contains 14 prespecified scenarios. The identifiable baseline is
the main  setting and satisfies the required causal assumptions by
construction. The remaining scenarios modify specific aspects of the problem, including need--care alignment and mismatch, unmet need, over-intensive care, risk--benefit alignment and misalignment, absence of shared response patterns,
strong observed confounding, hidden confounding, poor overlap, null and placebo outcomes, capacity scarcity, and heterogeneous intervention costs.

The scenarios are analyzed separately rather than averaged into a single score
because they test different properties of the framework. Poor-overlap and hidden-confounding settings examine failures of causal identification. Null and placebo settings test whether the system abstains when no beneficial response exists. Capacity-scarcity and heterogeneous-cost settings preserve the causal
response structure while modifying the operational allocation problem.


\paragraph{ASL Benevento Semi-Synthetic Study}
\label{sec:asl-data}

The ASL study is constructed from pseudonymized administrative data describing
270,533 unique assisted individuals. The available sources include the
assisted-population registry, hospitalization records, diagnosis records, and
exemption records. These sources are linked to construct one patient-level
record containing demographic characteristics, exemption status, prior
hospital use, morbidity proxies, frailty proxies, and indicators of
administrative care complexity.

The application index date is December~31, 2025. Only information recorded
before this date is included among the learner-facing covariates. After data
cleaning and application of the age, activity, and temporal requirements,
230,525 adults were available for sampling. Each experimental run samples
30,000 individuals from this population.

The real covariate matrix is combined with three prespecified semi-synthetic
response scenarios. In the \emph{partially aligned} scenario, prognostic risk
and incremental benefit share part of their structure, while additional
profile-specific heterogeneity prevents them from inducing the same ordering.
In the \emph{risk--benefit misaligned} scenario, patients with greater
prognostic risk are not necessarily those expected to obtain the largest
incremental benefit, testing whether the framework can distinguish prediction
of adverse outcomes from causal prioritization. Finally, the
\emph{mixed-response} scenario introduces more heterogeneous profile-specific
responses, including a larger proportion of neutral and harmful effects.

All three scenarios preserve the same real covariate structure but generate
different treatment assignments, potential outcomes, and profile-specific
incremental benefits. They are evaluated separately because they represent
increasingly difficult relationships between prognostic risk and causal
response.

The partially aligned scenario is used as the main ASL reference setting. In
this scenario, an average of 22,176 patients per run were eligible for at least
one candidate intervention. Because one patient may be eligible for multiple interventions, these patients generated an average of 50,080 admissible patient--intervention opportunities. Table~\ref{tab:asl-cohort-flow} summarizes
the source data, the resulting analytic population, and the average cohort flow
across the ten partially aligned runs.

\begin{table}[h]
\centering
\footnotesize
\caption{ASL source and analytic cohort summary. Analytic counts are averages
over ten partially aligned runs.}
\label{tab:asl-cohort-flow}
\begin{tabular}{lr}
\toprule
Element & Count \\
\midrule
Unique assisted individuals & 270,533 \\
Hospitalization records & 20,825 \\
Diagnosis records & 49,057 \\
Curated exemption records & 283,297 \\
Adults available before sampling & 230,525 \\
Patients sampled per run & 30,000 \\
Patients with at least one admissible intervention & 22,176 \\
Admissible patient--intervention opportunities & 50,080 \\
\bottomrule
\end{tabular}
\end{table}

The ASL semi-synthetic design preserves the observed covariate distribution and
administrative heterogeneity, while treatment assignments, potential outcomes,
incremental benefits, and oracle decisions are generated after the covariate
matrix is fixed. The three response mechanisms represent partial risk--benefit alignment,
explicit risk--benefit misalignment, and a mixed-response setting with more
neutral and harmful profile-specific effects. Oracle results refer exclusively to the
semi-synthetic data-generating process and must not be interpreted as estimates
of the actual effects of ASL Benevento services. Additional data-quality and
response diagnostics are reported in
Appendix~\ref{app:asl-data-quality}.

\subsection{Evaluation Metrics}
\label{sec:partitions-metrics}

Metrics follow the four decision layers of the framework. Baseline-need
recovery is evaluated through \emph{weighted Cohen's $\kappa$},
\emph{ordinal mean absolute error}, and \emph{within-one-level accuracy}.
These quantities respectively measure ordinal agreement, the average distance
from the reference level, and the proportion of predictions differing from the
reference by at most one level.

Ranking quality is measured through concordance. \emph{Held-out doubly robust
concordance} compares the learned score ordering with pseudo-outcomes computed
on held-out patients and is the primary non-oracle metric. \emph{Oracle
concordance}, available only in the synthetic and semi-synthetic settings after
model freezing, compares the ranking with the known generated benefits.
\emph{Within-intervention concordance}, \emph{cross-intervention concordance},
and \emph{global concordance} evaluate different portions of the common
priority ordering.

Recommendation metrics include the \emph{positive recommendation rate},
\emph{oracle recommendation agreement}, \emph{recommendation regret},
\emph{abstention rate}, and \emph{selected harm rate}. \emph{Within-patient
accuracy} is additionally reported in the ASL study.

Allocation performance is summarized by the \emph{number of patients served},
\emph{conditional deferral}, \emph{allocation regret}, and
\emph{normalized allocated value}, which compares the benefit obtained by the
model with the oracle value under the same constraints. \emph{AUTOC} and
\emph{Value@$k\%$} are used for the restricted-capacity baseline comparison.
Stability across repeated fits is evaluated through \emph{rank-score Spearman
correlation} and \emph{Jaccard overlap} of recommended or top-ranked
opportunity sets. Complete definitions are reported in
Appendix~\ref{app:experimental-details}.

\section{Results}
\label{sec:results}
We first evaluate the complete pipeline in the  synthetic setting,
then examine selected stress scenarios, the ASL semi-synthetic study, contrastive pretraining, and stability. Held-out doubly robust metrics use only
observationally available signals, whereas oracle quantities are accessed only after the non-oracle pipeline has been frozen.

\paragraph{Synthetic Performance}
\label{sec:confirmatory-results}
\begin{table}[h]
\centering
\scriptsize
\caption{Synthetic results over ten  seeds. Values are
means with 95\% run-level bootstrap confidence intervals.}
\label{tab:main-results}
\begin{tabular}{lll}
\toprule
Decision layer & Metric & Estimate [95\% CI] \\
\midrule
Need
& Weighted Cohen's $\kappa$
& 0.705 [0.699--0.710] \\
Need
& Ordinal MAE
& 0.696 [0.691--0.702] \\
Need
& Within-one-level accuracy
& 0.899 [0.896--0.902] \\
\midrule
Ranking
& Held-out DR within-intervention concordance
& 0.502 [0.494--0.509] \\
Ranking
& Held-out DR cross-intervention concordance
& 0.546 [0.534--0.561] \\
Ranking
& Held-out DR global concordance
& 0.538 [0.527--0.550] \\
Ranking
& Oracle cross-intervention concordance
& 0.832 [0.816--0.849] \\
Ranking
& Oracle global concordance
& 0.799 [0.781--0.817] \\
\midrule
Recommendation
& Recommendation rate
& 0.524 [0.497--0.549] \\
Recommendation
& Oracle-intervention agreement
& 0.815 [0.791--0.842] \\
Recommendation
& Mean recommendation regret
& 0.168 [0.107--0.228] days \\
\midrule
Allocation
& Conditional deferral rate
& 0.361 [0.325--0.397] \\
Allocation
& Normalized allocated value
& 0.936 [0.915--0.953] \\
Allocation
& Allocation regret
& 549.2 [404.7--723.3] days \\
\bottomrule
\end{tabular}
\end{table}

Table~\ref{tab:main-results} reports the  results in the synthetic setting. The need module achieved a weighted Cohen's $\kappa$ of 0.705, with 89.9\% of predictions within one level of the generated reference.
Oracle cross-intervention and global concordance reached 0.832 and 0.799,
respectively, showing that the model recovered a common ordering across
interventions. The lower held-out DR concordance reflects the observation-level noise of doubly robust pseudo-outcomes.
A positive intervention was recommended for 52.4\% of patients, with 81.5\%
agreement with the oracle-best intervention. Under capacity constraints, the
allocation retained 93.6\% of oracle value, although 36.1\% of positively
recommended patients were deferred because the required capacity was
unavailable.

\paragraph{Scenario-Level Robustness Analysis}
\label{sec:scenario-results}

\begin{table}[h]
\centering
\footnotesize
\caption{Selected synthetic stress scenarios. Support and abstention are
proportions; normalized value compares the model allocation with the oracle
allocation under the same constraints.}
\label{tab:scenario-main}
\resizebox{\columnwidth}{!}{%
\begin{tabular}{lcccc}
\toprule
Scenario
& Support
& Abstention
& Oracle global concordance
& Normalized value \\
\midrule
Poor overlap
& 0.798
& 0.630
& 0.734
& 0.949 \\
Hidden confounding
& 1.000
& 0.462
& 0.567
& 0.893 \\
Risk--benefit misaligned
& 1.000
& 0.490
& 0.681
& 0.975 \\
Capacity scarcity
& 1.000
& 0.462
& 0.809
& 0.885 \\
Sharp null
& 1.000
& 1.000
& --
& 1.000 \\
Placebo outcome
& 1.000
& 1.000
& --
& 1.000 \\
\bottomrule
\end{tabular}%
}
\end{table}

Table~\ref{tab:scenario-main} summarizes the main stress scenarios, while
complete results are reported in Appendix~\ref{app:scenario-diagnostics}.
Poor overlap reduced empirical support and increased abstention, whereas hidden
confounding produced the largest causal degradation.
Under risk--benefit misalignment, the model retained most of the oracle
allocation value despite weaker correspondence between prognostic risk and
incremental benefit. Capacity scarcity instead reduced allocation value while
leaving ranking quality largely unchanged, separating operational constraints
from ranking failure.
Sharp-null and placebo scenarios produced complete abstention because no
meaningful positive treatment ordering was present. Overall, the scenarios
distinguish failures caused by limited support, invalid causal identification,
and restricted service capacity.

\paragraph{ASL Semi-Synthetic Results}
\label{sec:asl-results}

\paragraph{ASL Semi-Synthetic Results}
\label{sec:asl-results}

\begin{table*}[t]
\centering
\scriptsize
\setlength{\tabcolsep}{6pt}
\renewcommand{\arraystretch}{1.15}
\caption{ASL semi-synthetic results over ten runs per scenario. Values are
means with 95\% run-level bootstrap confidence intervals. Oracle quantities
refer only to the generated response mechanism.}
\label{tab:asl-main-results}
\begin{tabular}{lcccccc}
\toprule
Scenario
& \shortstack{Held-out DR\\global}
& \shortstack{Oracle\\global}
& \shortstack{Oracle\\cross-profile}
& \shortstack{Within-patient\\accuracy}
& \shortstack{Normalized\\value}
& \shortstack{Selected\\harm} \\
\midrule
Partially aligned
& 0.625 [0.616--0.633]
& 0.766 [0.759--0.773]
& 0.801 [0.793--0.809]
& 0.800 [0.783--0.816]
& 0.838 [0.828--0.847]
& 0.001 [0.000--0.002] \\

Risk--benefit misaligned
& 0.616 [0.606--0.627]
& 0.776 [0.770--0.781]
& 0.812 [0.805--0.818]
& 0.788 [0.778--0.799]
& 0.846 [0.827--0.865]
& 0.006 [0.004--0.009] \\

Mixed response
& 0.578 [0.570--0.586]
& 0.648 [0.642--0.653]
& 0.673 [0.665--0.681]
& 0.646 [0.636--0.656]
& 0.660 [0.645--0.677]
& 0.119 [0.112--0.126] \\
\bottomrule
\end{tabular}
\end{table*}

Table~\ref{tab:asl-main-results} reports the three ASL response mechanisms. In
the partially aligned and risk--benefit-misaligned scenarios, oracle global
concordance remained above 0.76, normalized allocation value was approximately
0.84, and selected harm remained below 1\%.
The mixed-response scenario was substantially harder: oracle global concordance
decreased to 0.648, within-patient accuracy to 0.646, and normalized allocation
value to 0.660, while selected harm increased to 11.9\%. Since approximately
900 patients were served in every scenario, this reduction reflects poorer
prioritization rather than fewer available service slots.
Held-out DR concordance remained lower than oracle concordance because of
pseudo-outcome noise. Oracle results refer only to the generated
semi-synthetic mechanism and not to the actual effects of ASL Benevento
services.

\paragraph{Contrastive Pretraining and Stability}
\label{sec:contrastive-results}
\begin{table}[t]
\centering
\footnotesize
\caption{Change in normalized allocation value from contrastive pretraining
relative to random initialization.}
\label{tab:contrastive-main}
\begin{tabular}{lc}
\toprule
ASL scenario & Difference [95\% CI] \\
\midrule
Partially aligned
& $+0.0132$ [0.0016, 0.0243] \\
Risk--benefit misaligned
& $+0.0155$ [0.0046, 0.0283] \\
Mixed response
& $+0.0210$ [0.0099, 0.0326] \\
\bottomrule
\end{tabular}
\end{table}

As shown in Table~\ref{tab:contrastive-main}, contrastive pretraining improved
normalized allocation value in all ASL scenarios, with the largest gain in the
mixed-response setting.
Across repeated fits on fixed synthetic data, rank-score Spearman correlation
was 0.738. Recommendation sets were highly stable, with a Jaccard similarity of
0.923, whereas top-10\% opportunity overlap was lower at 0.608, indicating
greater sensitivity near sharp ranking cutoffs. Complete ablation and stability
results are reported in Appendix~\ref{app:supplementary-results}.

\section{Discussion, Limitations, and Governance}
\label{sec:discussion}

The results support three main design choices. First, baseline need,
recommendation, and allocation must remain separate. Need describes clinical
and care complexity, recommendation identifies the supported intervention with
the largest expected incremental benefit, and allocation determines whether it
can be activated under current resources. Operational deferral must therefore
not be interpreted as low need, ineligibility, or absence of expected benefit.

Second, the relevant prioritization unit is the patient--intervention
opportunity rather than the patient alone, since one patient may be eligible
for interventions with different benefits, costs, and resource requirements.
Third, global ranking is meaningful only across interventions sharing the same
reference condition, outcome, follow-up horizon, measurement unit, and
direction of benefit. Interventions pursuing different objectives must remain
in separate decision domains.

The evidence remains preclinical. Synthetic findings depend on the selected
data-generating mechanisms, while the ASL study combines real administrative
covariates with simulated assignments and outcomes. It therefore neither
estimates the actual effects of DM77-aligned services nor establishes improved
clinical outcomes.

Observational application would require credible temporal alignment,
consistency, conditional exchangeability, positivity, and an appropriate
interference assumption. These conditions cannot be fully verified, and
unmeasured severity, professional judgment, social support, geography, and
service availability may affect both assignment and outcome, as illustrated by
the hidden-confounding scenario.

The current formulation represents a single planning cycle and does not model
repeated interventions, time-varying confounding, adherence, delayed effects,
service congestion, or dynamic capacity. Calibration, support thresholds,
costs, and population shift may also alter recommendations and allocation
value. The contribution is therefore the formulation and evaluation of a
governed global causal ranking problem.

Clinical eligibility must remain external to the learned ranking.
\texttt{PROMETHEUS} must not govern access to urgent, essential, mandatory,
palliative, or legally protected care. A score cannot create eligibility,
remove a contraindication, or override professional judgment. Since historical
data may reflect unequal access, incomplete documentation, geography, and
socioeconomic disadvantage, deployment would require subgroup audits of
eligibility, empirical support, abstention, recommendation, harm, deferral, and
allocation.

Decision records should distinguish baseline need, admissible interventions,
the recommended and allocated interventions, empirical-support diagnostics,
and the reason for abstention or operational deferral. Clinicians must retain
authority to accept, modify, postpone, or reject every recommendation.

\section{Deployment Status and Operational Pathway}
\label{sec:deployment}

\texttt{PROMETHEUS} is currently implemented as a preclinical decision-support
prototype rather than as a live clinical system. The workflow supports
patient-level data preparation, eligibility evaluation, causal supervision,
global ranking, recommendation, and capacity-constrained allocation. Its
outputs explicitly distinguish baseline need, recommended intervention,
allocated intervention, abstention, and operational deferral.

The ASL Benevento study evaluates the technical feasibility of applying this
workflow to heterogeneous administrative data, but no generated recommendation
has been used to modify patient care. A controlled deployment would first
operate in silent mode, allowing healthcare professionals to inspect
recommendations without affecting clinical decisions. Evaluation would include
data-processing reliability, empirical-support coverage, recommendation
stability, clinician agreement, override reasons, subgroup disparities,
allocation feasibility, and system latency. Progression beyond this stage
would require institutional authorization, prospective validation, continuous
monitoring, and explicit clinician responsibility for every final decision.

\section{Conclusion}
\label{sec:conclusion}

\texttt{PROMETHEUS} formulates population-health planning as a global causal
ranking of comparable patient--intervention opportunities, followed by
patient-level recommendation and capacity-constrained allocation. Controlled
experiments show promising recovery of generated priorities and allocation
value, while also exposing sensitivity to poor overlap, hidden confounding,
and heterogeneous harmful responses. Prospective use requires observational
validation, explicit governance, subgroup auditing, and clinician oversight.

\section*{Ethics and Privacy Statement}

The ASL study uses pseudonymized administrative covariates and simulated
assignments and outcomes. \texttt{PROMETHEUS} is intended as decision support,
not as an autonomous gatekeeper for care. Any application requires
privacy-preserving data governance, transparent abstention and deferral
reasons, fairness auditing, and professional authority over final decisions.

\section*{Acknowledgments}

This work was conducted within the institutional collaboration between the
Department of Computer Science, University of Salerno, and ASL Benevento.

\bibliographystyle{ACM-Reference-Format}
\bibliography{sample-base}

\appendix

\section{Technical Formalization}
\label{app:technical-details}

This appendix provides the formal definitions underlying
Section~\ref{sec:problem-formulation}. It focuses on comparability,
identification, recommendation, and allocation; conceptual motivation is kept
in the main text.

\subsection{Comparability and Identification}

For two patient--profile opportunities $u=(i,p)$ and $v=(j,q)$, define
\[
M_{uv}^{\mathrm{comp}}
=
\begin{cases}
1, & \text{if $p$ and $q$ belong to the same comparability domain},\\
0, & \text{otherwise}.
\end{cases}
\]
Same-profile comparisons are automatically comparable. Within-patient and
cross-profile comparisons are used only when
$M_{uv}^{\mathrm{comp}}=1$; profiles pursuing different clinical objectives
remain in separate ranking domains.

For profile $p$, the causal contrast is defined in
\[
\mathcal I_p
=
\left\{
i:
E_i^{(p)}=1,\;
A_i\in\{0,p\}
\right\},
\qquad
D_i^{(p)}=\mathbb I(A_i=p),
\]
with pre-decision adjustment variables
\[
W_i
=
\left(
X_i,\,
C_i,\,
\widehat L_i^{\mathrm{need}}
\right).
\]
Identification of $\tau_p(w)$ requires consistency, conditional
exchangeability,
\[
\left\{
Y_i^{(p)},Y_i^{(0)}
\right\}
\perp
D_i^{(p)}
\mid
W_i,\,
E_i^{(p)}=1,
\]
positivity,
\[
0
<
e_p(w)
=
\Pr
\left(
D_i^{(p)}=1
\mid
W_i=w,\,
E_i^{(p)}=1
\right)
<
1,
\]
and an application-appropriate interference assumption. These conditions are
profile-specific. Clinical admissibility does not itself guarantee empirical
support: an admissible opportunity may still be excluded from training or
automatic recommendation when overlap, treatment prevalence, effective sample
size, or repeated-estimation stability is inadequate.

\subsection{Recommendation and Allocation}

Let $Q_i^{(p)}$ indicate empirical support and define
$\mathcal P_i^{\mathrm{sup}}
=\{p\in\mathcal P_i^{\mathrm{adm}}:Q_i^{(p)}=1\}$.
For a nonempty supported set,
$p_i^\star=\arg\max_{p\in\mathcal P_i^{\mathrm{sup}}}\widehat b_{i,p}$.
Given a minimum-benefit threshold $\delta$,
\[
\widehat A_i^{\mathrm{rec}}
=
\begin{cases}
p_i^\star, & \widehat b_{i,p_i^\star}>\delta,\\
0, & \widehat b_{i,p_i^\star}\leq\delta,\\
\bot, & \mathcal P_i^{\mathrm{sup}}=\varnothing,
\end{cases}
\]
where $0$ denotes maintenance and $\bot$ denotes abstention.

The binary allocation variables $Z_{i,p}$ maximize
\[
\sum_i\sum_{p\in\mathcal P_i^{\mathrm{sup}}}
Z_{i,p}\widehat b_{i,p}
\]
subject to
\[
\begin{aligned}
Z_{i,p}
&\leq
E_i^{(p)}Q_i^{(p)}
\mathbb I\{\widehat b_{i,p}>\delta\},\\
\sum_p Z_{i,p}
&\leq 1,\\
\sum_{i,p}c_{i,p,r}Z_{i,p}
&\leq K_r,
\qquad r\in\mathcal R.
\end{aligned}
\]
The resource index $r$ can represent profile capacity, shared workforce,
budget, geography, prerequisites, or workload limits. A positive recommendation
that cannot be activated is recorded as an operational deferral.

\section{Training, Reliability, and Support Details}
\label{app:training-details}

Profile-specific cross-fitting is repeated $R$ times. The repeated supervision
signal and nuisance-instability estimate are
\[
\overline\Gamma_i^{(p)}
=
R^{-1}\sum_{r=1}^R\Gamma_{i,r}^{(p)},
\qquad
\sigma_i^{(p)}
=
\operatorname{sd}_r(\Gamma_{i,r}^{(p)}).
\]
Under the assumptions in Appendix~\ref{app:technical-details},
$\mathbb E[\Gamma_i^{(p)}\mid W_i=w,E_i^{(p)}=1]=\tau_p(w)$.
This equality holds in conditional expectation; an individual pseudo-outcome
is not an observed individual effect.

For a retained pair $u=(i,p)$ and $v=(j,q)$, the reliability weight is
\[
\begin{aligned}
\omega_{uv}
={}&
\omega_{uv}^{\mathrm{sep}}
\omega_{uv}^{\mathrm{stab}}\\
&\times
\omega_{uv}^{\mathrm{overlap}}
\omega_{uv}^{\mathrm{ESS}}.
\end{aligned}
\]
The four terms account for causal-signal separation, repeated-estimation
stability, propensity overlap, and profile-level effective sample size.
Weights are normalized within comparison type so that within-profile,
within-patient, and cross-profile pairs contribute on controlled scales.

Support filtering remains distinct from clinical eligibility:
\[
\begin{aligned}
Q_i^{(p)}
={}&
Q_{i,p}^{\mathrm{overlap}}
Q_{i,p}^{\mathrm{stability}}\\
&\times
Q_p^{\mathrm{ESS}}
Q_p^{\mathrm{prevalence}}.
\end{aligned}
\]
All thresholds, pair proportions, calibration choices, and the
minimum-benefit threshold are selected using non-oracle validation data and
frozen before confirmation. Training pairs contain ranker-training patients
only; validation, calibration, and test pseudo-outcomes never enter pair
construction.

\section{Experimental Protocol Details}
\label{app:experimental-details}

The synthetic benchmark contains 14 prespecified scenarios with 5,000 patients
each. Patients are partitioned without overlap into nuisance training (30\%),
ranker training (32.5\%), validation/calibration (25\%), and testing (12.5\%).
Profile-specific nuisance models use three-fold cross-fitting repeated five
times. Potential outcomes, true effects, true propensities, latent variables,
and oracle decisions are stored in separate evaluation artifacts.

Six discretionary profiles share the outcome ``days alive outside acute
hospital care over 365 days.'' The protected palliative pathway uses a
different objective and is excluded from the discretionary global ranking.
Five discovery seeds are used for non-oracle component selection, followed by
ten untouched confirmation seeds and five fresh populations for the final
baseline benchmark. Confidence intervals are computed by run-level bootstrap,
and model comparisons use matched seeds and populations.

The primary non-oracle ranking metric is held-out DR global concordance.
Oracle quantities are opened only after model freezing. Recommendation,
allocation, harm, restricted-capacity, and stability metrics are defined in
Section~\ref{sec:partitions-metrics}; this appendix reports the protocol details
needed to reproduce them.

\section{Supplementary Results}
\label{app:supplementary-results}

This appendix reports complete scenario-level diagnostics and secondary
analyses omitted from the main text for space. It includes robustness across
all prespecified synthetic settings, restricted-capacity performance,
contrastive ablations, stability, and implementation checks.

\subsection{Synthetic Scenario Diagnostics}
\label{app:scenario-diagnostics}

Table~\ref{tab:all-scenario-diagnostics} extends the selected stress tests in
Table~\ref{tab:scenario-main} to all 14 frozen synthetic scenarios.

\begin{table*}[t]
\centering
\small
\caption{Synthetic scenario diagnostics for the 14 prespecified scenarios. Each row is
one frozen 5,000-patient realization. Oracle concordance is undefined under
tied null effects.}
\label{tab:all-scenario-diagnostics}
\begin{tabular}{lrrrrr}
\toprule
Scenario & Support & Abstention & Oracle cross-profile & Oracle global & Norm. value \\
\midrule
Baseline identifiable & 1.000 & 0.566 & 0.841 & 0.808 & 0.964 \\
Need--care alignment & 0.993 & 0.402 & 0.840 & 0.794 & 0.945 \\
Unmet need & 1.000 & 0.099 & 0.905 & 0.871 & 0.887 \\
Over-intensive care & 0.864 & 0.907 & 0.363 & 0.538 & 1.000 \\
Risk--benefit aligned & 1.000 & 0.558 & 0.816 & 0.783 & 0.918 \\
Risk--benefit misaligned & 1.000 & 0.490 & 0.680 & 0.681 & 0.975 \\
No shared response & 1.000 & 0.549 & 0.811 & 0.772 & 0.972 \\
Strong observed confounding & 1.000 & 0.667 & 0.811 & 0.779 & 0.960 \\
Poor overlap & 0.798 & 0.630 & 0.788 & 0.734 & 0.949 \\
Sharp null & 1.000 & 1.000 & -- & -- & 1.000 \\
Placebo outcome & 1.000 & 1.000 & -- & -- & 1.000 \\
Hidden confounding & 1.000 & 0.462 & 0.557 & 0.567 & 0.893 \\
Capacity scarcity & 1.000 & 0.462 & 0.845 & 0.809 & 0.885 \\
Heterogeneous costs & 1.000 & 0.457 & 0.847 & 0.821 & 0.946 \\
\bottomrule
\end{tabular}
\end{table*}

The complete results separate causal, statistical, and operational failure
modes. Hidden confounding yields the largest ranking degradation, while poor
overlap reduces support and increases abstention. The scenarios not emphasized
in the main text show that unmet need remains highly rankable but is harder to
serve under the available resources, whereas over-intensive care produces high
abstention and weak ranking recovery. Its normalized value of one should
therefore be read conditionally on the small set of retained opportunities.
No-shared-response and strong-observed-confounding settings remain
substantially more recoverable than hidden confounding, and heterogeneous
costs primarily alter allocation rather than the learned causal ordering.

\subsection{Restricted-Capacity, Ablation, and Stability Results}
\label{app:secondary-results}

Table~\ref{tab:compact-supplementary-results} collects secondary quantities not
shown in the main result tables.

\begin{table*}[t]
\centering
\caption{Compact supplementary results not shown in the main tables.}
\label{tab:compact-supplementary-results}
\small
\setlength{\tabcolsep}{4pt}
\begin{tabular}{llc}
\toprule
Analysis & Metric or comparison & Estimate [95\% CI] \\
\midrule
Restricted capacity & PROMETHEUS AUTOC & 1.7487 \\
Restricted capacity & PROMETHEUS Value@10\% / Value@20\% & 0.8361 / 1.4981 \\
Joint contrastive ablation & Rank-only minus rank+contrastive, held-out DR & $+0.000241$ [$-0.000455$, $+0.000985$] \\
Joint contrastive ablation & Rank-only minus rank+contrastive, oracle & $-0.001284$ [$-0.003551$, $+0.000727$] \\
Stability & Rank-score Spearman & 0.738 [0.666--0.797] \\
Stability & Recommended-patient Jaccard & 0.923 [0.905--0.941] \\
Stability & Top-10\% Jaccard & 0.608 [0.529--0.689] \\
Diagnostics & Prespecified gates passed & 8/8 \\
\bottomrule
\end{tabular}
\end{table*}

The restricted-capacity metrics confirm that the learned ordering retains
useful value when only a limited fraction of opportunities can be selected.
Both joint-contrastive confidence intervals include zero, supporting the
sequential use of contrastive learning for initialization rather than as a
continuing auxiliary objective. Recommendation sets are more stable than
membership near the sharp top-10\% cutoff, as shown by the larger
recommended-patient Jaccard value. All eight prespecified gates were passed,
and no feasibility violation occurred across the 14 canonical allocations.

\section{ASL Data Quality and Supplementary Diagnostics}
\label{app:asl-data-quality}

This appendix documents cohort construction, source quality, profile
prevalence, and supplementary ASL results. As stated in
Section~\ref{sec:asl-data}, the administrative covariates are observed, whereas
assignments, potential outcomes, incremental benefits, and oracle decisions are
generated by the locked semi-synthetic mechanism. Oracle quantities therefore
evaluate generated targets and are not estimates of the effects of services
delivered by ASL Benevento.

\subsection{Source Tables and Record Linkage}
\label{app:asl-source-linkage}

Table~\ref{tab:asl-source-tables} expands the source counts summarized in the
main paper and records the registry deduplication performed before
patient-level feature construction.

\begin{table}[t]
\centering
\caption{Dimensions of the ASL administrative sources before
application-specific sampling and semi-synthetic generation.}
\label{tab:asl-source-tables}
\begin{tabular}{lr}
\toprule
Source element & Count \\
\midrule
Raw assisted-population registry rows & 275,397 \\
Unique assisted individuals & 270,533 \\
Registry identifiers appearing more than once & 4,864 \\
Curated exemption records & 283,297 \\
Hospitalization records & 20,825 \\
Diagnosis records in long format & 49,057 \\
Patient-feature records & 270,533 \\
\bottomrule
\end{tabular}
\end{table}

The registry contained 4,864 repeated pseudonymous identifiers, corresponding
to 9,728 rows. Deterministic preprocessing resolved these duplicates without
detecting age disagreements greater than one year. Two exact duplicate
exemption rows were removed before constructing the curated table.

Table~\ref{tab:asl-linkage-diagnostics} reports cross-source linkage and
integrity checks.

\begin{table}[t]
\centering
\caption{Record-linkage and integrity diagnostics for the ASL source
tables.}
\label{tab:asl-linkage-diagnostics}
\begin{tabular}{lr}
\toprule
Diagnostic & Value \\
\midrule
Linkable hospitalization rows & 20,819 \\
Unlinkable hospitalization rows & 6 \\
Unique patients in hospitalization data & 17,949 \\
Hospitalized patients found in registry & 9,339 \\
Hospitalization-to-registry linkage rate & 52.0\% \\
Unique patients in exemption data & 190,491 \\
Exemption patients found in registry & 187,277 \\
Exemption-to-registry linkage rate & 98.3\% \\
Post-index death patients & 1,465 \\
\bottomrule
\end{tabular}
\end{table}

Hospitalization linkage was materially lower than exemption linkage, reflecting
different temporal and organizational coverage across the supplied sources.
The hospitalization-to-registry rate must therefore not be interpreted as the
population prevalence of hospitalization. Post-index events were excluded from
learner-facing features. No hospitalization had a negative calendar-derived
length of stay; 3,008 records with a discrepancy greater than one day were
retained with an audit flag. Exemption codes potentially affected by
scientific-notation conversion were processed as strings and preserved with
their quality indicators.

\subsection{Feature Missingness}
\label{app:asl-missingness}

Table~\ref{tab:asl-patient-missingness} summarizes the principal patient-level
missingness patterns after preprocessing.

\begin{table}[t]
\centering
\caption{Selected missingness diagnostics for the patient-level registry.
Variables with structurally meaningful absence are marked separately.}
\label{tab:asl-patient-missingness}
\begin{tabular}{lr}
\toprule
Variable & Missing proportion \\
\midrule
Patient identifier & 0.0\% \\
Sex & 0.0\% \\
Age at index date & 0.0\% \\
Residence municipality & $<0.01$\% \\
Enrollment district & 0.0\% \\
Provider district & 0.0\% \\
Provider code & 25.8\% \\
AFT code & 12.6\% \\
AFT name & 12.6\% \\
Provider type & 0.0\% \\
Income-exemption indicator & 0.0\% \\
Invalidity-exemption indicator & 0.0\% \\
Pathology-exemption indicator & 0.0\% \\
Recorded pre-index death year$^\dagger$ & 98.8\% \\
\bottomrule
\multicolumn{2}{l}{\footnotesize $^\dagger$Structurally absent for patients
without a recorded pre-index death.}
\end{tabular}
\end{table}

Missingness is concentrated in provider descriptors rather than in the main
demographic, exemption, and derived morbidity variables. The absence of a
recorded pre-index death year is structural for patients without a recorded
death and is represented through a separate indicator. In the hospitalization
extract, procedure description, type-of-bill code, and the original principal
diagnosis field were not treated as universally available features; fields
entirely unavailable in the supplied source were excluded. No post-index
information was used to impute pre-decision covariates.

\subsection{Baseline Covariate Distribution}
\label{app:asl-baseline}

As reported in Section~\ref{sec:asl-data}, each run sampled 30,000 patients from
230,525 eligible active adults. Table~\ref{tab:asl-baseline-characteristics}
provides the corresponding baseline distributions for the partially aligned
reference scenario.

\begin{table}[t]
\centering
\caption{Selected baseline characteristics of the sampled ASL population in
the partially aligned reference scenario. Values are averages across ten
runs.}
\label{tab:asl-baseline-characteristics}
\begin{tabular}{lr}
\toprule
Characteristic & Estimate \\
\midrule
Patients sampled per run & 30,000 \\
Mean age, years & 52.14 \\
Female patients & 50.96\% \\
Income-exemption indicator & 66.21\% \\
Invalidity-exemption indicator & 9.89\% \\
Pathology-exemption indicator & 22.11\% \\
Mean previous hospitalization count & 0.040 \\
Mean morbidity proxy & 1.033 \\
Mean frailty proxy & 0.189 \\
\bottomrule
\end{tabular}
\end{table}

Baseline distributions were stable across the ten runs and the three response
mechanisms because the same sampling protocol was used throughout. The low mean
hospitalization count reflects both the rarity of linked inpatient events in
the supplied extract and incomplete registry overlap; it is not a complete
regional or lifetime hospitalization rate. All learner-facing variables were
constructed from information available before the index date.

\subsection{Profile Eligibility and Assignment Prevalence}
\label{app:asl-profile-diagnostics}

The main paper reports the aggregate number of eligible patients and admissible
opportunities. Table~\ref{tab:asl-profile-population} provides the corresponding
profile-level eligibility, simulated assignment, and generated response
diagnostics.

\begin{table*}[t]
\centering
\caption{Profile-level eligibility, simulated assignment, and oracle-only
response diagnostics. Values are averages across ten runs. True-effect
quantities are generated evaluation targets, not observed ASL treatment
effects.}
\label{tab:asl-profile-population}
\begin{tabular}{llrrrr}
\toprule
Scenario & Profile & Eligibility & Assignment & Mean true benefit & Negative effects \\
\midrule
Partially aligned
& Profile 1 & 72.18\% & 16.38\% & 1.62 & 7.05\% \\
& Profile 2 & 22.76\% & 5.88\%  & 1.72 & 6.62\% \\
& Profile 3 & 72.00\% & 20.37\% & 2.90 & 6.20\% \\
\midrule
Mixed response
& Profile 1 & -- & -- & -- & 20.39\% \\
& Profile 2 & -- & -- & -- & 16.94\% \\
& Profile 3 & -- & -- & -- & 18.92\% \\
\bottomrule
\end{tabular}
\end{table*}

Eligibility and simulated assignment are strongly imbalanced across profiles,
supporting evaluation in a realistic multi-treatment setting rather than a
balanced experimental matrix. The mixed-response mechanism increases the
fraction of negative effects to approximately 17--20\%, which explains the
larger harm and lower allocation value observed in that scenario.

In the current ASL implementation, eligibility defines the complete
patient--profile opportunity set. Positivity is monitored through stabilized
propensities, minimum arm sizes, and profile-level effective-sample-size
diagnostics, but no additional opportunity-level exclusion is applied after
eligibility. Consequently, supported and admissible opportunity counts are
identical in this application. This implementation-specific choice does not
alter the general distinction between eligibility and empirical support.

\subsection{Supplementary ASL Evaluation and Ablation Results}
\label{app:asl-complete-results}

Primary ranking, within-patient, selected-harm, and normalized-allocation
results are reported in Table~\ref{tab:asl-main-results}.
Table~\ref{tab:asl-selected-complete} reports only the additional quantities
omitted from the main text.

\begin{table*}[t]
\centering
\caption{Additional ASL semi-synthetic metrics omitted from the main result
table. Values are means with 95\% run-level bootstrap confidence intervals.}
\label{tab:asl-selected-complete}
\begin{tabular}{lccc}
\toprule
Metric & Partially aligned & Risk--benefit misaligned & Mixed response \\
\midrule
NDCG@25\% & 0.917 [0.912--0.921] & -- & -- \\
Harm@25\% & 0.0008 [0.0004--0.0014] & -- & -- \\
Patients served & 911.3 [886.9--935.5] & 901.1 [870.0--930.9] & 910.7 [883.1--934.8] \\
Allocation regret & 936.7 [884.4--995.1] & -- & -- \\
\bottomrule
\end{tabular}
\end{table*}

The similar number of patients served across scenarios confirms that the lower
mixed-response allocation value is driven by prioritization quality rather than
by a materially different number of service slots. NDCG@25\% and Harm@25\%
provide additional top-region diagnostics for the partially aligned reference
setting.

\subsubsection{Contrastive-Pretraining Ablation}

Table~\ref{tab:asl-ablation-complete} reports paired differences between the
selected contrastively pretrained configuration and all principal ablations.
Positive values favor the selected model.

\begin{table*}[t]
\centering
\caption{Paired normalized-allocation-value differences over ten ASL runs
per scenario. Positive differences favor the contrastively pretrained
configuration.}
\label{tab:asl-ablation-complete}
\small
\setlength{\tabcolsep}{4pt}
\resizebox{\textwidth}{!}{%
\begin{tabular}{llcc}
\toprule
Scenario & Comparator & Oracle difference [95\% CI] & Non-oracle validation difference [95\% CI] \\
\midrule
Partially aligned
& Random initialization
& $+0.0132$ [$+0.0016$, $+0.0243$]
& $+0.0032$ [$+0.0014$, $+0.0051$] \\

& Joint UACT
& $+0.0133$ [$+0.0017$, $+0.0246$]
& $+0.0032$ [$+0.0014$, $+0.0052$] \\

& Legacy triplet
& $+0.0106$ [$+0.0010$, $+0.0197$]
& $+0.0033$ [$+0.0013$, $+0.0056$] \\

& Pretrained and fine-tuned
& $0.0000$ [$0.0000$, $0.0000$]
& $0.0000$ [$-0.0000$, $0.0000$] \\
\midrule
Risk--benefit misaligned
& Random initialization
& $+0.0155$ [$+0.0046$, $+0.0283$]
& $+0.0013$ [$-0.0021$, $+0.0042$] \\

& Joint UACT
& $+0.0156$ [$+0.0048$, $+0.0286$]
& $+0.0013$ [$-0.0021$, $+0.0042$] \\

& Legacy triplet
& $+0.0102$ [$-0.0028$, $+0.0207$]
& $+0.0024$ [$-0.0005$, $+0.0058$] \\

& Pretrained and fine-tuned
& $-0.0000$ [$-0.0000$, $0.0000$]
& $-0.0000$ [$-0.0000$, $0.0000$] \\
\midrule
Mixed response
& Random initialization
& $+0.0210$ [$+0.0099$, $+0.0326$]
& $+0.0083$ [$+0.0032$, $+0.0135$] \\

& Joint UACT
& $+0.0207$ [$+0.0089$, $+0.0334$]
& $+0.0084$ [$+0.0033$, $+0.0133$] \\

& Legacy triplet
& $+0.0215$ [$+0.0064$, $+0.0374$]
& $+0.0074$ [$+0.0024$, $+0.0123$] \\

& Pretrained and fine-tuned
& $+0.0001$ [$0.0000$, $+0.0003$]
& $0.0000$ [$-0.0000$, $0.0000$] \\
\bottomrule
\end{tabular}%
}
\end{table*}

The full table adds two conclusions beyond the compact main-text comparison.
First, the gain over joint UACT is similar to the gain over random
initialization, indicating that merely retaining a contrastive term during
ranking does not reproduce the benefit of pretraining. Second, the selected
model generally improves over the legacy triplet construction, although the
risk--benefit-misaligned oracle interval includes zero. The comparison with the
pretrained-and-fine-tuned variant is effectively zero, supporting removal of
the unnecessary contrastive retention loss from the final configuration.

\subsubsection{Interpretive Limits}

The locked aggregate artifacts contain ranking, harm, patients-served, and
allocation-value metrics. Exact recommendation counts, profile-specific
recommendation distributions, and explicit oracle-profile agreement require
opportunity-level score artifacts that were not produced by the original runs.
These quantities are therefore not reconstructed retrospectively.
Within-patient accuracy retains its original metric name, and patients served
is not used as a substitute for the number of positive pre-allocation
recommendations.

\section{Implementation Details and Hyperparameter Configuration}
\label{app:network-details}

This section reports the frozen ASL implementation. Learner hyperparameters
affect estimation, protocol parameters define the experimental design, and
operational parameters define the allocation environment. Model selection used
only non-oracle validation signals.

\subsection{Semi-Synthetic Protocol and Data Splits}

Table~\ref{tab:asl-protocol-hyperparameters} reports the sampling, partitioning,
seed, and run configuration used for reproducibility.

\begin{table*}[htbp]
\centering
\caption{Semi-synthetic study and reproducibility settings. The smoke-test configuration was used only for engineering diagnostics and not for the reported confirmation results.}
\label{tab:asl-protocol-hyperparameters}
\small
\setlength{\tabcolsep}{4pt}
\begin{tabular}{p{0.25\textwidth}p{0.13\textwidth}p{0.56\textwidth}}
\toprule
Setting & Value & Interpretation \\
\midrule
Main sample size & 30,000 & Number of adult ASL patients used in a complete experimental run. \\
Smoke-test sample size & 6,000 & Reduced sample used only to verify execution and diagnostic outputs. \\
Minimum age & 18 years & Patients younger than 18 years were excluded before sampling. \\
Default response scenario & \nolinkurl{partially_aligned} & Default semi-synthetic relationship between baseline risk and profile-specific benefit. Campaign experiments additionally considered \nolinkurl{risk_benefit_misaligned} and \nolinkurl{mixed_response}. \\
Patient-level split proportions & 30/30/15/10/15\% & Proportions assigned respectively to nuisance training, ranking training, validation, calibration, and testing. \\
Data-generation seed & 8101 & Controls patient sampling, simulated treatment assignment, potential outcomes, and patient-level partitioning. \\
Nuisance seed & 8102 & Controls repeated nuisance-model fitting and stratified bootstrap samples. \\
Ranker seed & 8103 & Controls neural-network initialization, pair ordering, dropout streams, and contrastive sampling. \\
Training-pair seed & 8104 & Controls construction of the ranking pairs in the ranking-training partition. \\
Validation-pair seed & 8105 & Controls construction of the held-out validation pairs. \\
Discovery runs & 5 per scenario & Paired runs used for non-oracle model development and configuration decisions. \\
Ablation runs & 5 per scenario & Paired runs comparing the contribution of pretraining, uncertainty handling, calibration, and within-patient supervision. \\
Confirmation runs & 10 per scenario & Untouched runs performed after freezing the selected configuration. \\
\bottomrule
\end{tabular}
\end{table*}

All opportunities belonging to the same patient inherit the patient-level
partition, preventing leakage across nuisance estimation, ranking, validation,
calibration, and testing. Numeric variables are standardized using
training-partition statistics, and categorical variables are one-hot encoded
with unseen categories ignored at scoring time.

\subsection{Nuisance Estimation and Doubly Robust Supervision}

Table~\ref{tab:nuisance-hyperparameters} reports the active nuisance-model and
repeated-DR settings.

\begin{table*}[htbp]
\centering
\caption{Hyperparameters used to estimate the repeated doubly robust supervision signals.}
\label{tab:nuisance-hyperparameters}
\small
\setlength{\tabcolsep}{4pt}
\begin{tabular}{p{0.25\textwidth}p{0.13\textwidth}p{0.56\textwidth}}
\toprule
Hyperparameter & Value & Role \\
\midrule
Nuisance fitting mode & \texttt{full\_bootstrap} & Fits every nuisance repetition on a treatment-stratified bootstrap sample of the nuisance-training partition. \\
Number of nuisance repetitions & 3 & Number of independently fitted propensity and outcome-model sets. The resulting repeated DR contrasts are used to measure sign agreement and uncertainty. \\
Minimum observations per arm & 20 & Minimum number of nuisance-training observations required for each treatment arm. The run is rejected when this support condition is not satisfied. \\
Minimum stabilized propensity & 0.04 & Lower bound imposed on every multinomial treatment probability. Probabilities are shrunk toward the uniform distribution while preserving a unit row sum. \\
Propensity boosting iterations & 65 & Maximum number of iterations for the multinomial histogram gradient-boosting classifier. \\
Outcome boosting iterations & 70 & Maximum number of iterations for each treatment-specific histogram gradient-boosting outcome model. \\
Maximum leaf nodes & 15 & Maximum number of leaves in each propensity and outcome boosting tree. This is a fixed implementation value. \\
Nuisance boosting learning rate & 0.06 & Shrinkage parameter used by both the propensity and outcome boosting models. This is a fixed implementation value. \\
\bottomrule
\end{tabular}
\end{table*}

Each eligible opportunity receives three repeated DR signals generated from
treatment-stratified bootstrap nuisance fits. Their mean provides supervision,
while their dispersion and directional agreement support pair and triplet
reliability checks:
\[
\overline{\Gamma}_{i}^{(p)}
=
\frac{1}{3}\sum_{r=1}^{3}\Gamma_{i,r}^{(p)}.
\]

\subsection{Ranker Architecture and Optimization}

The model is a shared profile-conditioned Siamese network. Table~\ref{tab:ranker-hyperparameters}
reports its architecture and optimization settings.

\begin{table*}[htbp]
\centering
\caption{Architecture and ranking-optimization hyperparameters.}
\label{tab:ranker-hyperparameters}
\small
\setlength{\tabcolsep}{4pt}
\begin{tabular}{p{0.25\textwidth}p{0.13\textwidth}p{0.56\textwidth}}
\toprule
Hyperparameter & Value & Role \\
\midrule
Shared hidden width & 96 & Width of the clinical encoder, FiLM-conditioned representation, and post-conditioning layer. \\
Profile-embedding dimension & 16 & Dimension of the learned embedding associated with each candidate care profile. \\
Projection dimension & 24 & Dimension of the normalized representation used by the contrastive losses. \\
Dropout probability & 0.12 & Dropout applied in the clinical encoder and after FiLM conditioning. \\
Optimizer & AdamW & Optimizer used for both contrastive pretraining and ranking fine-tuning. \\
Global weight decay & $5\times10^{-4}$ & Weight decay used during ranking fine-tuning. \\
Encoder fine-tuning learning rate & $1\times10^{-4}$ & Learning rate applied to the pretrained encoder, treatment embedding, FiLM module, post-FiLM layer, and projection head. \\
Ranking-head learning rate & $5\times10^{-4}$ & Learning rate applied to the scalar score head. \\
Fallback learning rate & $3\times10^{-4}$ & Default learning rate used only when separate encoder and head rates are not provided. It is therefore not active in the frozen configuration. \\
Maximum fine-tuning epochs & 60 & Maximum number of ranking-training epochs. \\
Minimum fine-tuning epochs & 20 & Earliest point at which a fine-tuning checkpoint may be selected. \\
Early-stopping patience & 12 & Number of validation checks without improvement allowed after the earliest admissible checkpoint. \\
Ranking batch size & 256 & Number of ranking pairs processed in each primary optimization step. \\
Initial encoder freeze & 3 epochs & Only the score head is updated during the first three fine-tuning epochs, protecting the pretrained representation from abrupt changes. \\
Freeze encoder for entire fine-tuning & False & The encoder is unfrozen after the initial three epochs. \\
Gradient-norm clipping & 5.0 & Maximum Euclidean gradient norm used in both pretraining and fine-tuning. This is a fixed implementation value. \\
\bottomrule
\end{tabular}
\end{table*}

The clinical encoder has two layers of width 96, a 16-dimensional profile
embedding, FiLM conditioning, a scalar score head, and a normalized
24-dimensional contrastive projection. The score remains an ordinal priority
quantity rather than a calibrated individual treatment-effect estimate.

\subsection{Construction of Ranking Pairs}

Ranking pairs comprise within-patient, within-profile, and global cross-profile
comparisons. Table~\ref{tab:pair-hyperparameters} reports their sampling,
filtering, and weighting settings.

\begin{table*}[htbp]
\centering
\caption{Hyperparameters controlling ranking-pair construction and weighting.}
\label{tab:pair-hyperparameters}
\small
\setlength{\tabcolsep}{4pt}
\begin{tabular}{p{0.25\textwidth}p{0.13\textwidth}p{0.56\textwidth}}
\toprule
Hyperparameter & Value & Role \\
\midrule
Maximum training pairs & 30,000 & Maximum number of stable pairs sampled from the ranking-training partition. \\
Maximum validation pairs & 9,000 & Maximum number of independently sampled stable pairs used for held-out model selection. \\
Minimum DR-signal gap & 0.15 & Minimum absolute difference between the average repeated DR signals of the two opportunities. \\
Minimum repeat agreement & 0.70 & At least 70\% of nuisance repetitions must support the direction indicated by the average DR difference. \\
Within-patient pair fraction & 0.50 & Target proportion of comparisons involving different profiles for the same patient. \\
Within-profile pair fraction & 0.20 & Target proportion comparing different patients for the same profile. \\
Global cross-profile fraction & 0.30 & Target proportion comparing different patients and different profiles. \\
Top-region fraction & 0.25 & Defines the upper 25\% of opportunities according to the average DR signal. \\
Top-region pair multiplier & 2.0 & Multiplies the weight of any pair containing an opportunity in the upper benefit region. \\
Gap-clipping quantile & 0.95 & Clips the DR-gap component of the pair weight at its 95th percentile. \\
Within-patient loss multiplier & 1.75 & Increases the relative contribution of same-patient profile comparisons in the primary ranking loss. \\
Within-profile loss multiplier & 1.0 & Leaves the contribution of within-profile population comparisons unchanged. \\
Global cross-profile multiplier & 1.0 & Leaves the contribution of cross-profile population comparisons unchanged. \\
\bottomrule
\end{tabular}
\end{table*}

A pair is retained only when the repeated DR signals provide a sufficiently
large and stable ordering. Gap clipping, repeat agreement, top-region weighting,
and family-specific multipliers control the influence of each pair. Weights are
renormalized to mean one after applying these adjustments.

\subsection{Uncertainty-Aware Contrastive Pretraining}

The final pretraining implementation uses ordinal triplets rather than a
continuing contrastive loss during ranking. Let
\[
\widetilde{\Gamma}_{i,p}
=
\frac{\overline{\Gamma}_{i}^{(p)}}
{\operatorname{sd}(\overline{\Gamma})},
\]
where scaling without mean-centering preserves the interpretation of zero
incremental benefit. Reliable opportunities are grouped into negative,
neutral, and positive response regions. The encoder minimizes
\[
\mathcal L_{\mathrm{pre}}
=
\frac{
\sum_{(u,v,w)}
\rho_{uvw}
\left[
m_{\mathrm{pre}}
-
\operatorname{sim}(z_u,z_v)
+
\operatorname{sim}(z_u,z_w)
\right]_{+}
}{
\sum_{(u,v,w)}\rho_{uvw}
}.
\]
Table~\ref{tab:pretraining-hyperparameters} reports all active pretraining
settings.

\begin{table*}[htbp]
\centering
\caption{Hyperparameters of uncertainty-aware contrastive pretraining.}
\label{tab:pretraining-hyperparameters}
\small
\setlength{\tabcolsep}{4pt}
\begin{tabular}{p{0.25\textwidth}p{0.13\textwidth}p{0.56\textwidth}}
\toprule
Hyperparameter & Value & Role \\
\midrule
Pretraining enabled & True & Initializes the shared representation before fitting the ranking score. \\
Contrastive mode & \nolinkurl{uncertainty_aware_ordinal_triplet} & Uses reliable positive, neutral, and negative ordinal response neighborhoods. \\
Maximum pretraining epochs & 30 & Maximum number of encoder-only contrastive epochs. \\
Pretraining patience & 8 & Number of validation epochs without improvement before stopping pretraining. \\
Pretraining learning rate & $1\times10^{-3}$ & AdamW learning rate for the representation parameters during pretraining. \\
Pretraining weight decay & $5\times10^{-4}$ & Weight decay applied during contrastive pretraining. \\
Cosine triplet margin & 0.15 & Required separation between anchor--positive and anchor--negative cosine similarities. \\
Neutral-effect quantile & 0.20 & The data-dependent neutral threshold is initially defined by the 20th percentile of the absolute standardized DR signal. \\
Minimum neutral threshold & 0.05 & Lower bound on the standardized magnitude defining non-neutral response. \\
Positive-radius quantile & 0.20 & Defines the candidate positive radius from the lower 20\% of within-class effect distances. \\
Minimum positive radius & 0.05 & Lower bound on the standardized effect distance used for positive neighborhoods. \\
Negative-radius quantile & 0.70 & Defines clear negatives using the 70th percentile of distances among reliable opportunities. \\
Minimum negative radius & 0.25 & Lower bound on the standardized effect distance required for a clear negative. \\
Uncertainty filtering & True & Excludes opportunities whose dispersion across repeated DR signals exceeds the selected uncertainty cutoff. \\
Maximum uncertainty quantile & 0.60 & Retains opportunities at or below the 60th percentile of repeated-signal uncertainty. \\
Uncertainty weighting & True & Weights triplets according to the reliability of their constituent opportunities. \\
Reliability floor & 0.10 & Lower bound placed on the reliability score before triplet weights are constructed. \\
Maximum anchors per epoch & 12,000 & Maximum number of eligible anchors sampled in each pretraining epoch. Anchors are balanced across the three response classes when possible. \\
Anchor batch size & 128 & Number of candidate anchors processed in each pretraining optimization step. \\
Maximum validation anchors & 4,000 & Maximum number of held-out anchors used to assess the pretrained representation. \\
Maximum validation triplets & 6,000 & Maximum number of held-out triplets used to compute the pretraining selection criterion. \\
Same-patient hard negatives & False & Pretraining uses global clear negatives rather than forcing alternative profiles for the same patient to be negatives. \\
Same-patient negative probability & 0.0 & Probability of sampling a same-patient hard negative; inactive because same-patient negatives are disabled. \\
Strict diagnostics & True & Stops the run when no valid anchors, triplets, checkpoint, or non-zero encoder gradient is produced. \\
\bottomrule
\end{tabular}
\end{table*}

Checkpoint selection uses
\[
M_{\mathrm{pre}}
=
0.40\,\mathrm{TripletAcc}
+
0.30\,\mathrm{Separation}
+
0.30\,\mathrm{EffectDistCorr}.
\]
The criterion balances margin satisfaction, representation separation, and
agreement between latent and learner-safe causal-signal distances.

\subsection{Ranking Fine-Tuning Objectives}

After pretraining, the encoder and score head are optimized using
\[
\mathcal L_{\mathrm{total}}
=
\mathcal L_{\mathrm{rank}}
+
0.15\,\mathcal L_{\mathrm{within}}
+
0.10\,\mathcal L_{\mathrm{cal}}.
\]
The primary ranking loss remains dominant. The within-patient term reinforces
profile choice for the same patient, while the calibration regularizer
encourages a monotone relationship between scores and standardized learner-safe
DR signals. The selected final model does not retain a contrastive loss during
fine-tuning. Table~\ref{tab:finetuning-objective-hyperparameters} lists the
active auxiliary settings.

\begin{table*}[htbp]
\centering
\caption{Active auxiliary objectives used during ranking fine-tuning. The
contrastive retention loss was excluded from the selected final configuration.}
\label{tab:finetuning-objective-hyperparameters}
\small
\setlength{\tabcolsep}{4pt}
\begin{tabular}{p{0.27\textwidth}p{0.13\textwidth}p{0.54\textwidth}}
\toprule
Hyperparameter & Value & Role \\
\midrule
Calibration objective & Enabled & Adds a reliability-weighted pointwise loss between ranking scores and standardized learner-safe DR signals. \\
Calibration weight & 0.10 & Coefficient multiplying the calibration regularizer. \\
Calibration batch size & 256 & Number of opportunities used in each calibration update. \\
Huber transition parameter & 1.0 & Parameter $\beta$ of the smooth $L_1$ calibration loss. \\
Minimum calibration reliability & 0.25 & Lower bound on the reliability weight assigned to a calibration target. \\
Within-patient auxiliary & Enabled & Adds an additional pairwise objective over alternative profiles available to the same patient. \\
Within-patient weight & 0.15 & Coefficient multiplying the within-patient auxiliary loss. \\
Maximum within-patient pairs & 12,000 & Maximum number of additional same-patient comparisons. \\
Minimum within-patient signal gap & 0.05 & Minimum absolute difference between two profile-specific mean DR signals for the same patient. \\
Minimum within-patient agreement & 0.60 & Minimum fraction of nuisance repetitions agreeing on the comparison direction. \\
Within-patient batch size & 256 & Number of auxiliary same-patient pairs processed per update. \\
Contrastive retention during fine-tuning & Disabled & The final configuration uses contrastive learning for encoder initialization only. \\
\bottomrule
\end{tabular}
\end{table*}

\subsection{Held-Out Checkpoint Selection}

Fine-tuning checkpoints are selected with the non-oracle criterion
\[
\begin{split}
M_{\mathrm{rank}}
={}&
0.20\,C_{\mathrm{global}}
+
0.25\,C_{\mathrm{cross}}
+
0.30\,C_{\mathrm{within}}\\
&+
0.15\,\mathrm{NDCG}_{25\%}
+
0.10\,V_{\mathrm{top}\text{-}k}.
\end{split}
\]
Table~\ref{tab:validation-selection-hyperparameters} reports the component
weights and evaluation fractions.

\begin{table*}[htbp]
\centering
\caption{Hyperparameters of non-oracle checkpoint selection.}
\label{tab:validation-selection-hyperparameters}
\small
\setlength{\tabcolsep}{4pt}
\begin{tabular}{p{0.30\textwidth}p{0.13\textwidth}p{0.51\textwidth}}
\toprule
Validation component & Weight or value & Interpretation \\
\midrule
Global pair concordance & 0.20 & Agreement over all held-out ranking comparisons. \\
Cross-profile concordance & 0.25 & Agreement over pairs involving different patients and different profiles. \\
Within-patient concordance & 0.30 & Agreement when ranking alternative profiles for the same patient. \\
NDCG fraction & 0.25 & Computes NDCG over the upper 25\% of validation opportunities. \\
NDCG weight & 0.15 & Contribution of top-region ranking quality to checkpoint selection. \\
Top-$k$ DR-value fractions & 0.10, 0.25, 0.50 & Fractions used to evaluate the benefit captured by the highest-ranked opportunities. \\
Top-$k$ DR-value weight & 0.10 & Contribution of the average top-$k$ DR-value ratio. \\
Early-stopping tolerance & $10^{-7}$ & Minimum numerical increase required to reset patience. This is a fixed implementation value. \\
\bottomrule
\end{tabular}
\end{table*}

The earliest eligible checkpoint is epoch 20, which is more restrictive than
the initial encoder freeze and ensures that model selection is not driven by
the early transient phase of fine-tuning.

\subsection{Operational Allocation Parameters}

Table~\ref{tab:allocation-hyperparameters} reports the external capacity and
budget settings used after model training.

\begin{table*}[htbp]
\centering
\caption{Operational capacity and budget parameters.}
\label{tab:allocation-hyperparameters}
\small
\setlength{\tabcolsep}{4pt}
\begin{tabular}{p{0.30\textwidth}p{0.13\textwidth}p{0.51\textwidth}}
\toprule
Parameter & Value & Interpretation \\
\midrule
Monitoring capacity fraction & 0.18 & Monitoring capacity is set to 18\% of the number of test patients. \\
Home-care capacity fraction & 0.10 & Home-care capacity is set to 10\% of the number of test patients. \\
Case-management capacity fraction & 0.06 & Case-management capacity is set to 6\% of the number of test patients. \\
Budget multiplier & 0.85 & The available budget equals 85\% of the cost required to fill all profile-specific capacity pools. \\
\bottomrule
\end{tabular}
\end{table*}

These values define the degree of service scarcity; they do not affect
representation learning, pair construction, or checkpoint selection. The mixed
integer program assigns at most one profile per patient while respecting all
profile capacities and the total budget.

\subsection{Baseline Hyperparameters}

Table~\ref{tab:baseline-hyperparameters} reports the principal supervised
baseline settings used in the ASL experiments.

\begin{table*}[htbp]
\centering
\caption{Main baseline-model hyperparameters used in the full ASL experiments.}
\label{tab:baseline-hyperparameters}
\small
\setlength{\tabcolsep}{4pt}
\begin{tabular}{p{0.25\textwidth}p{0.20\textwidth}p{0.49\textwidth}}
\toprule
Baseline & Hyperparameters & Description \\
\midrule
Prospective-utilization Ridge & $\alpha=1.0$ & Ridge regression fitted to the simulated prospective-utilization outcome. \\
Prospective-utilization GBDT & 80 iterations; 15 leaf nodes & Histogram gradient-boosting risk model. \\
T-learner GBDT & 70 iterations per outcome model; 15 leaf nodes & Separate outcome regressions for the reference and candidate treatment arms. \\
X-learner GBDT & 60 outcome iterations; 50 pseudo-effect iterations; 50 propensity iterations; 15 leaf nodes & Profile-specific X-learners. Binary propensities are clipped to $[0.05,0.95]$. \\
DR-learner GBDT & 70 iterations; 15 leaf nodes & Profile-specific regressors fitted to the mean repeated DR signal. \\
DR Random Forest & 120 trees; minimum leaf size 12 & Profile-specific random-forest regressors fitted to the mean repeated DR signal. \\
PROMETHEUS post-hoc calibration & Isotonic regression; clipped extrapolation & A monotone calibrator is fitted on the separate calibration partition for policy-value comparisons. \\
\bottomrule
\end{tabular}
\end{table*}

The baseline specifications are reported for reproducibility and use the same
patient partitions and learner-facing covariates as PROMETHEUS. Oracle
quantities are not used for fitting or tuning any baseline.

\section{Inactive and Ablation-Only Configuration Entries}

The configuration retains backward-compatible entries that are inactive in the
selected model. The legacy response-binned triplet module is disabled and is
activated only by the \nolinkurl{legacy_triplet} ablation. Its stored parameters
include 6,000 maximum triplets, 10 quantile bins, positive and negative band
distances of 1 and 4, minimum repeat agreement 0.75, margin 0.50, and batch
size 256.

The top-level entries
\nolinkurl{maximum_contrastive_pairs=5000},
\nolinkurl{contrastive_weight=0.05},
\nolinkurl{contrastive_margin=1.0},
\nolinkurl{contrastive_warmup_epochs=1}, and
\nolinkurl{minimum_contrastive_epochs=2}
belong to the legacy implementation and do not affect the selected
uncertainty-aware ordinal-triplet pretraining procedure. The unused entries
\nolinkurl{contrastive_positive_quantile},
\nolinkurl{contrastive_negative_quantile}, and the legacy contrastive batch
size can be removed without changing the frozen output.

Supervised InfoNCE settings, including temperature 0.10 and hard-negative
weight 0.25, are used only in the corresponding ablation. Likewise, UACT
retention settings are preserved for the joint-contrastive ablation but are
disabled in the selected final configuration. Their presence in the
configuration file must therefore not be interpreted as evidence that a
contrastive auxiliary loss contributed to the reported final ranking model.

\end{document}